\documentclass[11pt,a4paper]{article}
\usepackage[hyperref]{acl2019}
\usepackage{times}
\usepackage{booktabs}       
\usepackage{amsfonts}
\usepackage{amsmath}
\usepackage{algorithm}
\usepackage{algorithmic}
\usepackage{url}
\usepackage{multirow}
\usepackage{graphics}
\usepackage{graphicx}
\usepackage{float}
\definecolor{Orange}{rgb}{1,0.5,0}

\def\aclpaperid{1659}

\aclfinalcopy 

\title{Scheduled DropHead: A Regularization Method for Transformer Models}

\author{
Wangchunshu Zhou$^{1}$~\thanks{\ \ This work was done during the first author's internship at Microsoft Research Asia.} ~~~ Tao Ge$^{2}$ ~~~ Ke Xu$^{1}$ ~~~ Furu Wei$^2$ ~~~ Ming Zhou$^2$\\
$^1$Beihang University, Beijing, China\\
$^2$Microsoft Research Asia, Beijing, China\\
{\tt zhouwangchunshu@buaa.edu.cn, kexu@nlsde.buaa.edu.cn}\\
{\tt \{tage, fuwei, mingzhou\}@microsoft.com}}

\date{}

\begin{document}
\maketitle
\begin{abstract}
We introduce DropHead, a structured dropout method specifically designed for regularizing the multi-head attention mechanism which is a key component of transformer. In contrast to the conventional dropout mechanism which randomly drops units or connections, DropHead drops entire attention heads during training to prevent the multi-head attention model from being dominated by a small portion of attention heads. It can help reduce the risk of overfitting and allow the models to better benefit from the multi-head attention. Given the interaction between multi-headedness and training dynamics, we further propose a novel dropout rate scheduler to adjust the dropout rate of DropHead throughout training, which results in a better regularization effect. Experimental results demonstrate that our proposed approach can improve transformer models by 0.9 BLEU score on WMT14 En-De translation task and around 1.0 accuracy for various text classification tasks.
\end{abstract}

\section{Introduction}
The transformer~\cite{vaswani2017attention} architecture has shown state-of-the-art performance across a variety of Natural Language Processing (NLP) tasks. One key architectural innovation of the transformer model is the multi-head attention mechanism (MHA) where attention is computed independently by multiple parallel attention heads. Intuitively, the MHA mechanism allows the model to jointly attend to information from different representation
subspaces at different positions.
In practice, however, some recent studies ~\cite{voita2019analyzing,michel2019sixteen} show that in most cases, the MHA mechanism is dominated by only a small portion of attention heads which contributes most to the output. The other attention heads can even be removed at test time without significantly impacting the performance, indicating that the MHA mechanism is far from being fully exploited.



\begin{figure}
    \centering
    \includegraphics[width=1.\linewidth]{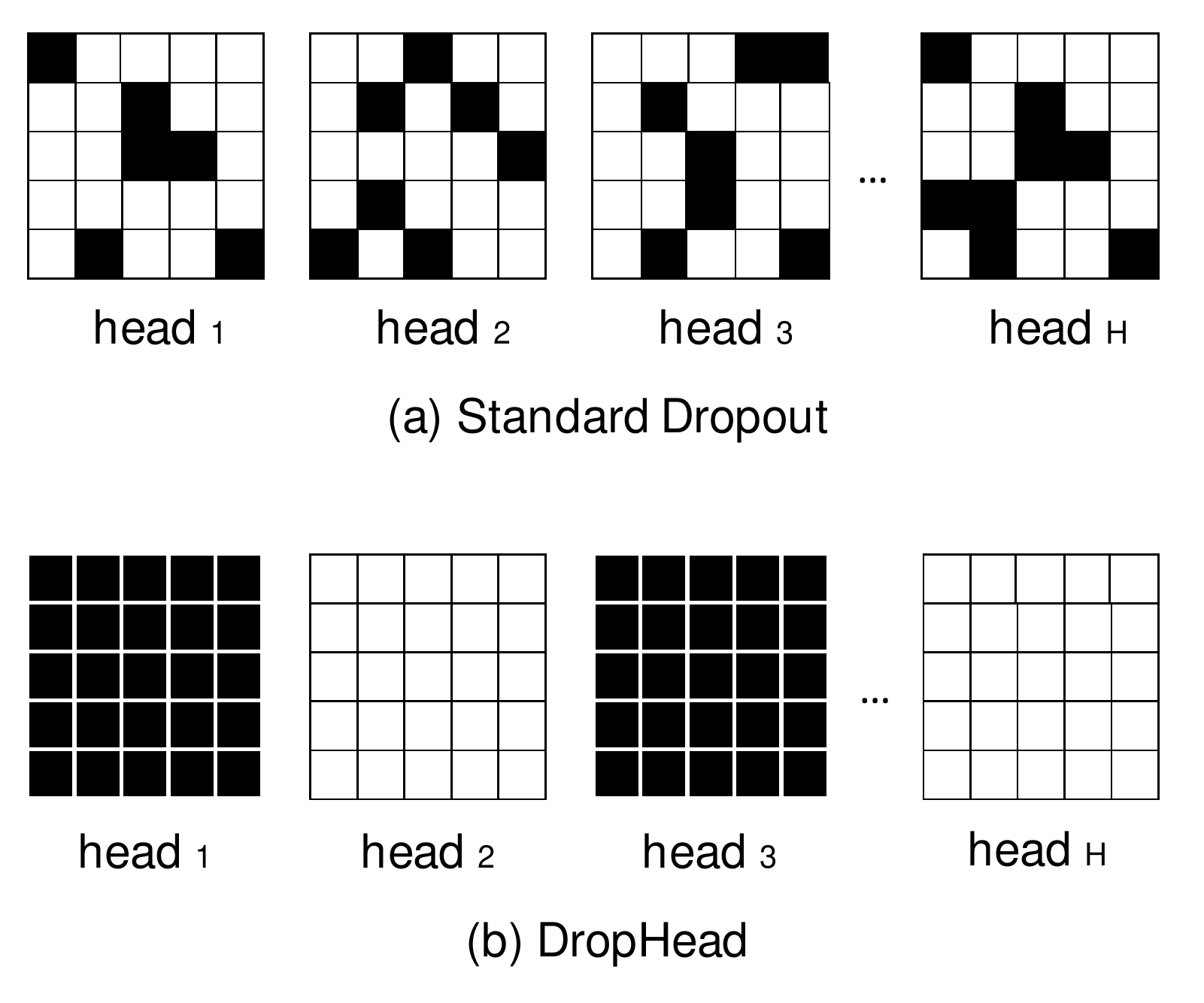}
    \caption{Illustration of DropHead : (a): standard Dropout randomly drops units in each attention heads. (b) DropHead drops entire attention heads.}
    \label{drophead}
\end{figure}

To address the limitation, we propose DropHead, a structured dropout mechanism specifically designed to regularize the MHA mechanism in transformer models. Different from conventional dropout approaches~\cite{srivastava2014dropout} that randomly dropout activations or connections, DropHead drops an entire attention head with a pre-defined probability, which is illustrated in Figure \ref{drophead}. Under the DropHead mechanism, any attention head is possible to be dropped out during training, which requires every attention head to work hard so that it can ensure the model to function well even if other heads are dropped out. As a result, DropHead 
can prevent the MHA from being dominated by one or a few attention heads and encourage more attention heads to encode useful information. Also, it helps MHA mechanism generalize better by breaking up excessive co-adaptation between multiple attention heads. 

Moreover, considering the training dynamics~\cite{michel2019sixteen} of the MHA where the aforementioned head domination phenomenon tends to happen at the very beginning of the training process, we propose a specific dropout rate scheduler for the DropHead mechanism, which looks like a V-shaped curve: it applies a relatively high dropout rate for the beginning and ending epochs during training to prevent the MHA mechanism from being dominated by a few attentions heads and reduce overfitting, while uses a lower dropout rate in the remaining steps to allow the model to better fit the training data.  

We conduct experiments on both neural machine translation and text classification benchmark datasets. Experimental results show that the proposed Scheduled DropHead yields consistent and substantial improvements over the standard transformer model (0.9 BLEU score on WMT2014 En-De translation dataset and around 1.0 accuracy on various text classification datasets) without any other modification in the model architecture. In addition, we also find that Scheduled DropHead enables a trained transformer model to better retain its performance after pruning out a relatively large number of attention heads.

Our contribution can be summarized as follows:
\begin{itemize}
    \item We propose DropHead, a simple and versatile regularization method for transformer models. It helps avoid the head domination problem, which enables more attention heads to encode useful information and prevents excessive co-adaptation between attention heads, improving the generalization ability of the model.
    \item We introduce a dropout scheduler for DropHead, which adjusts the dropout rate throughout the training process to strengthen the effect of DropHead, which further improves the performance.
\end{itemize}

\section{Background}

In this section, we briefly recall the important background notations including attention mechanism~\cite{bahdanau2014neural}, multi-head attention mechanism~\cite{vaswani2017attention}, and dropout~\cite{srivastava2014dropout}.

\paragraph{Attention Mechanism} An attention function can be described as mapping a query and a set of key-value pairs to an output, where the query, keys, values, and output are all vectors. The output is computed as a weighted sum of the values, where the weight assigned to each value is computed by a compatibility function of the query with the corresponding key. We focus on the dot-product attention~\cite{luong2015effective} which is employed in the transformer architecture.  

Given a sequence of vectors $H \in \mathbb{R}^{l\times d}$, where $l$ and $d$ represent the length of and the dimension of the input sequence, the self-attention projects $H$ into three different matrices: the query matrix $Q$, the key matrix $K$ and the value matrix vector $V$, and uses scaled dot-product attention to get the output representation:
\begin{equation}
\begin{array}{l}{Q, K, V=H W^{Q}, H W^{K}, H W^{V}} \\ {\operatorname{Attn}(Q, K, V)=\operatorname{softmax}\left(\frac{Q K^{T}}{\sqrt{d_{k}}}\right) V}\end{array}
\end{equation}
where $W^{Q}, W^{K}, W^{V}$ are learnable parameters and $\operatorname{softmax}()$ is performed row-wise.

\paragraph{Multi-Head Attention} To enhance the ability of self-attention, multi-head self-attention is introduced as an extension of the single head self-attention, which jointly model the multiple interactions from different representation spaces:

\begin{equation}
\begin{array}{l}{\text {MHA}(H)=\left[\text {head}_{1} ; \ldots ; \text {head}_{H}\right] W^{O},} \\ {\text { where head }_{i}=\operatorname{Attn}\left(H W_{i}^{Q}, H W_{i}^{K}, H W_{i}^{V}\right)}\end{array}
\end{equation}
where $W^{O}, W_{i}^{Q}, W_{i}^{K}, W_{}^{V}$ are learnable parameters. The transformer architecture is a stack of several multi-head self-attention layers and fully-connected layers. The multi-head attention mechanism extends the standard attention mechanism~\cite{bahdanau2014neural} by computing attention with H independent attention heads, thus allows the model to jointly attend to information from different representation subspaces at different positions. 

\paragraph{Dropout} Since introduced by~\citet{srivastava2014dropout}, dropout is extensively used in training deep neural networks. The method randomly suppresses neurons during training by setting it to 0 with probability $p$, which is called the dropout rate. The resulting model at test time is often interpreted as an average of multiple models during training, and it is argued that this can reduce overfitting and improve test performance. Dropout achieves great success on fully-connected layers. However, vanilla dropout fails to improve the performance of convolutional neural networks or recurrent neural networks and is only applied in the feed-forward layers in the transformer model, which motivates our proposed method.

\section{Scheduled DropHead}

In this section, we begin by introducing the proposed DropHead mechanism in detail and discuss its potential effects on training transformer models. Afterwards, we describe the proposed dropout rate scheduler specifically designed for DropHead.

\subsection{DropHead}

DropHead is a simple regularization method similar to Dropout~\cite{srivastava2014dropout} and DropConnect~\cite{wan2013regularization}. The main difference between DropHead and conventional Dropout is that we drop an entire attention head with a pre-defined probability in DropHead, instead of dropping individual units in conventional Dropout. 

Formally, during training, for a transformer layer with $H$ attention heads, we randomly sample a binary mask vector $\xi \in \{0,1\}^{H}$ with each element independently sampled from $\xi_{i} \sim \operatorname{Bernoulli}(p)$ where $p$ is the dropout rate. Afterward, we create $H$ attention head masks M, which is of the same shape with the output of one attention head, by filling all its elements with the corresponding $\xi_{i}$. We then apply the mask vectors $\text{M}_{i}$ to the corresponding output of attention heads $\text{head}_{i}$ by element-wise multiplication. Finally, we normalize the output of MHA to ensure the scale of the output representation matches between training and inference. The output of MHA mechanism equipped with DropHead can be formulated as:
\begin{equation}
\small
\begin{array}{l}{\text {MHA}(H)=\left([\text {head}_{1} \odot \text{M}_{i}; \ldots ; \text {head}_{H} \odot \text{M}_{H}\right] W^{O})/ \gamma,} \\ {\text {where head}_{i}=\operatorname{Attn}\left(H W_{i}^{Q}, H W_{i}^{K}, H W_{i}^{V}\right);} \\ {\ \ \ \ \ \ \ \ \ \ \ \ \ \ \ \ \ \gamma = \operatorname{sum}(\xi)/H}\end{array}
\end{equation}
where $\gamma$ is the rescale factor used for normalization and $\odot$ denotes element-wise product. The procedure of DropHead is summarized in Algorithm \ref{algo}.

\begin{algorithm}[tb]
   \caption{\textbf{DropHead}}
   \label{algo}
\begin{algorithmic}[1]
   \REQUIRE Dropout rate $p$; outputs from $H$ attention heads $O = [\text {head}_{1} ; \ldots ; \text {head}_{H}]$; mode
   \IF{mode $==$ inference}
   \STATE return O
   \ENDIF
   \STATE Randomly sample $\xi$: $\xi_{i} \sim \operatorname{Bernoulli}(p)$.
   \STATE For each element $\xi_{i}$, create a mask vector $\text {M}_{i}$ of the same shape as $\text {head}_{i}$ with all elements equal to $\xi_{i}$.
   \STATE Apply the mask: $O = O \times M$
   \STATE Normalization: $O = O \times H/\textbf{count\_ones}(\xi)$
\end{algorithmic}
\end{algorithm}

\paragraph{Effect of DropHead} Applying DropHead during training forces the MHA mechanism to optimize the same objective with different randomly selected subsets of its attention heads. Its potential effects can be summarized  as follows: It prevents the MHA from being dominated by few attention heads and reduces excessive co-adaptation between different attention heads, which helps MHA exploit its multi-headness more effectively and reduces the risk of overfitting, thus improving the generalization ability of trained models.

\subsection{Dropout rate scheduler}

\begin{figure}
    \centering
    \includegraphics[width=1.\linewidth]{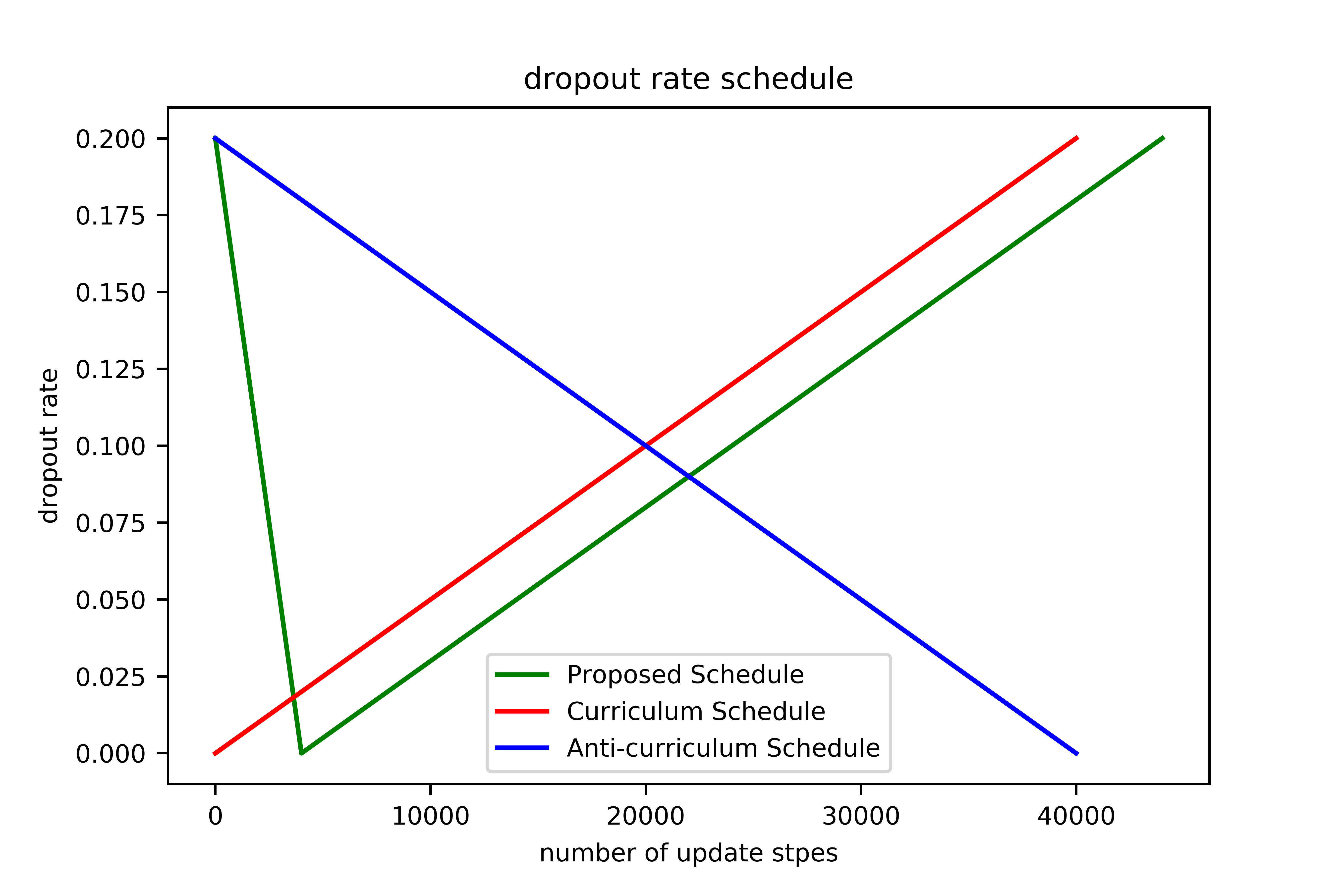}
    \caption{Illustration of the proposed dropout rate schedule, the curriculum dropout rate schedule, and the anti-curriculum dropout rate schedule.}
    \label{dropoutrate}
\end{figure}

Motivated by Curriculum Dropout~\cite{morerio2017curriculum} and other structured dropout approaches~\cite{ghiasi2018dropblock,zoph2018learning}, we propose to apply a dropout rate scheduler to adjust the effect of DropHead throughtout the training process. Different from the previous scheduled dropout approaches that generally initialize the training procedure with a dropout rate of 0 and gradually increase the dropout rate for alleviating the excessive co-adaptation in the late training steps, we propose a novel V-shaped dropout rate scheduler for DropHead, as shown in Figure \ref{dropoutrate} which additionally applies high dropout rate at the early stage of training. The motivation of high dropout rate at the beginning of training is to prevent the MHA from being dominated by few attention heads, which proves to happen at the very early stage of training, as the previous study~\cite{michel2019sixteen} shows.

Specifically, we start the training process with a relatively high dropout rate of $p_{start}$ and linearly decrease it to 0 during the early stage of training, which is empirically chosen to be the same training steps for learning rate warmup. Afterwards, we linearly increase the dropout rate to $p_{end}$ in the remaining training steps following existing dropout rate schedulers in DropBlock and ScheduledDropPath. To avoid introducing additional hyperparameters, we empirically set $p_{start} = p_{end}$ and find it generally works well across different tasks. 

\section{Experiments}

In this section, we first conduct experiments on several benchmark datasets on machine translation and text classification to demonstrate the effectiveness of Scheduled DropHead. Afterward, we conduct experiments on pruning transformer models trained with DropHead to test if it succeeds in preventing dominating attention heads and improving attention head pruning.

\subsection{Machine Translation}

We first conduct experiments on the machine translation task, which is one of the primary applications of the transformer model.

\paragraph{Dataset} Following~\citet{vaswani2017attention}, we use the WMT 2014 English-German dataset consisting of about 4.5 million sentence pairs as the training set. The newstest2013 and newstest2014 are used as the development set and the test set. We apply BPE~\cite{sennrich2015neural} with 32K merge operations to obtain subword units.

\paragraph{Models} We train the \textbf{transformer-big}~\cite{vaswani2017attention} model with the proposed Scheduled DropHead method as well as its counterpart where no dropout rate scheduler is employed. To ensure fair comparisons, we compare our models against the result reported by~\citet{vaswani2017attention} and our reproduced result. In addition, as our approach is expected to be able to reduce excessive co-adaptation between multiple attention heads, we also experiment with a variant of the transformer model where we increase the number of attention heads in each layer from 16 to 32. We denote the modified transformer model as \textbf{transformer-big (more heads)}. Note that we change the number of transformer heads by reducing the hidden dimension size of each attention heads, thus do not increase the number of total parameters in the resulting model. For reference, we also compare our model against that trained by applying the vanilla dropout on self-attention layers in the transformer model, as well as several recent techniques improving the transformer model in Table \ref{wmt14} to futher demonstrate the effectiveness of our approach.

\paragraph{Training} We generally follow the hyperparameter setting provided in~\cite{vaswani2017attention}. We set the dropout rate for DropHead to 0.2  and decrease the dropout rate in feed-forward layers from 0.3 to 0.2 based on dev set performance, which is consistent with the fact that DropHead mechanism already provides some regularization effects.

\paragraph{Evaluation} Following previous work, we average the last 10 checkpoints for evaluation. All results use beam search with a beam width of 4 and length penalty
of 0.6.

\begin{table}[!ht]
	\centering
	\resizebox{1.\linewidth}{!}{
		\begin{tabular}{lll}
			\toprule
			\bf Models & \bf BLEU & \bf PPL \\
			\midrule
			\textbf{Weighted Transformer}\cite{ahmed2017weighted} &  28.9  & -\\
			\textbf{Tied Transformer}\cite{xia2019tied} &  29.0  & - \\
			\textbf{Layer-wise Coordination}\cite{he2018layer} &  29.1  & -\\
			\midrule
			\textbf{Transformer-big} &  28.4  & - \\
			~- ours (reproduced) &  28.7  &  4.32 \\
			~- + Attention dropout & 28.7   &  4.29\\
			~- + DropHead & 29.2  &  4.15 \\
			~- + Scheduled DropHead & 29.4  & 4.08 \\
			\midrule
			\textbf{Transformer-big (more heads)} &  28.4 & 4.39 \\
			~- + Attention dropout & 28.5 & 4.38 \\
			~- + DropHead &  29.3 & 4.12 \\
			~- + Scheduled DropHead & \bf 29.6{*} & \bf 4.02 \\
			\bottomrule
	\end{tabular}}
	\caption{Machine translation performance on WMT14 en-de and newstest2014 of compared models. The results in the first section are the results reported on the corresponding publications. Other results are reported as the median of 5 random runs. {*} denotes statistically significant with $p<0.05$ compared with all baselines. }
	\label{wmt14}
\end{table}

\paragraph{Results}

Table \ref{wmt14} shows the BLEU score and the perplexity of different compared models. We find that applying vanilla dropout on self-attention layers generally fails to improve the model's performance. In contrast, the DropHead mechanism is able to improve the transformer model by around 0.5 BLEU score and 0.2 perplexity, demonstrating its ability to improve the model's generalization ability. In addition, the proposed dropout rate scheduler can further improve the performance of the strong baseline of Transformer + DropHead by 0.2 BLEU score, which confirms its effectiveness.

We also find that simply increasing the number of attention heads in each layer yields slightly worse result compared with the default model configuration. However, training the transformer model with more attention heads with Scheduled DropHead can yield further improvement compared with the original model architecture. This suggests that the proposed method is able to provide good regularization effect for the MHA in transformer. Our approach also substantially outperforms several existing techniques on improving the transformer model that do not involve large modification of the model architecture and additional training data, which further demonstrates its effectiveness.

\subsection{Text Classification}

We also conduct experiments on sentence classification tasks to further demonstrate the effectiveness of the proposed methods.

\begin{table}[!htt]
	\centering
	\scalebox{0.8}{
		\begin{tabular}{llcc}
			\toprule
			\bf Datasets & \bf Type & \bf \#Classes & \bf \#Documents \\
			\midrule
			\bf IMDB & Sentiment & 2 & 50000 \\
			\bf Yelp Review & Sentiment & 2 & 598000  \\
			\bf AG's News & Topic & 4 & 127,600 \\
			\bf DBPedia & Topic & 14 & 630,000 \\
			\bf TREC & Question & 6 & 5,952 \\
			\bf Yahoo! Answers & Question & 10 & 1,146,000 \\
			\bf SNLI & Inference & 3 & 550,152 \\
			\bottomrule
	\end{tabular}}
	\caption{Statistics for text classification datasets.}
	\label{data}
\end{table}

\paragraph{Datasets} We conduct experiments on text classification datasets ranging
from different tasks. Statistics of datasets are listed in Table 2. All datasets are split into training, development and testing sets following previous work. For sentiment analysis,
we use the binary film review \textbf{IMDB} dataset~\cite{maas2011learning} and the binary version of the \textbf{Yelp Review} dataset built by~\cite{zhang2015character}. For topic classification task, we employ \textbf{AG’s News} and \textbf{DBPedia} created by~\cite{zhang2015character}. For question classification task, we evaluate our method on the six-class version of the \textbf{TREC} dataset~\cite{voorhees1999trec} and \textbf{Yahoo! Answers} dataset created by~\cite{zhang2015character}. We also evaluate on the \textbf{SNLI} dataset~\cite{bowman2015large} which is a collection of sentence pairs labeled for entailment, contradiction, and semantic independence.

\begin{table*}[!ht]
	\centering
		\resizebox{1.\linewidth}{!}{
		\begin{tabular}{lccccccc}
			\toprule
			\bf Model & \bf IMDB & \bf Yelp & \bf AG & \bf DBPedia & \bf TREC & \bf Yahoo! & \bf SNLI \\
			\midrule
			\textbf{Char-level CNN}\cite{zhang2015character} & - & 95.12 & 90.49 & 98.45 & - & 71.20 & - \\
			\textbf{VDCNN}\cite{conneau2017supervised} & - & 95.72 & 91.33 & 98.71 & - & 73.43 & - \\
			\textbf{DPCNN}\cite{johnson2017deep} & - & 97.36 & 93.13 & 99.12 & - & 76.10 & - \\
			\textbf{ULMFiT}\cite{howard2018universal} & 95.40 & 97.84 & 94.99 & 99.20 & 96.40 & - & - \\
			\midrule
			\textbf{BERT-base} \cite{devlin2018bert} & 94.60 & 97.61 & 94.75 & 99.30 & 97.20 & 77.58 & 90.73 \\
			~-Attention Dropout  & 94.51 & 97.64 & 94.65 & 99.28 & 97.20 & 77.64 & 90.65 \\
			~-DropHead  & 95.22 & 97.77 & 94.90 & 99.35 & \bf 97.60 & \bf 78.05 & 90.85 \\
			~-Scheduled DropHead & \bf 95.48{*} & \bf \underline{97.92} & \bf 94.95 & \bf \underline{99.41} & \bf \underline{97.80} & 77.93 & \bf \underline{90.92} \\
			\midrule
			\textbf{Transformer (w/o pretraining)} & 90.85 & 95.61 & 91.08 & 98.69 & 94.60 & 73.15 & 86.52  \\
			~-Attention Dropout  & 90.91 & 95.56 & 91.12 & 98.62 & 94.40 & 73.31 & 86.57  \\
			~-DropHead & 91.87 & 96.14 & 91.68 & 98.85 & \bf 95.20 & 74.38 & 87.14  \\
			~-Scheduled DropHead & \bf 91.95{*} & \bf 96.21{*} & \bf 91.74{*} & \bf 98.91 & \bf 95.60{*} & \bf 74.41{*} & \bf 87.20{*} \\
			\bottomrule
	\end{tabular}}
	\caption{Test set results (accuracy) of compared models on text classification benchmarks. The results in the first section are the results reported on the corresponding publications. Other results are reported as the median of 5 random runs. ‘–’ means not reported results. {*} and \_  means statistically significant improvement upon the corresponding baseline with $p<0.05$ and $p< 0.1$.}
	\label{class}
\end{table*}

\paragraph{Models} 


We employ BERT-base~\cite{devlin2018bert} as a strong baseline and compare the performance of fine-tuning it without attention dropout, with vanilla attention dropout, and with DropHead/Scheduled DropHead. However, the dominant attention heads in the BERT model may  already be formed during pretraining, which may hinder the effect of DropHead, and pretraining BERT with our approach is beyond the scope of our paper. Therefore, we also employ a smaller, randomly initialized (i.e. without pretraining) transformer-based text classification model as our baseline model. The model has a hidden size of 768, 6 transformer blocks and 12 self-attention heads. Following~\cite{sun2019fine}, we train the models with a batch size of 24, a dropout rate of 0.1 for both vanilla Dropout and DropHead. We employ the Adam optimizer with a base learning rate as 2e-5 or 1e-4 with a warm-up ratio of 0.1 for the models with and without pretraining respectively. We train the models for 10 epochs and select the best model according to the accuracy on the held out dev set.


\paragraph{Results} We report the accuracy of compared models on the test set in Table \ref{class}. We can see that applying Scheduled DropHead can significantly (with $p<0.05$) improve the performance of vanilla transformer-based text classification models, as well as that applying vanilla dropout on the MHA mechanism of transformer models, on 6 out of 7 datasets. This demonstrates that our proposed approach works well for natural language understanding tasks. Scheduled DropHead can also consistently improve the performance of a very competitive BERT-based model with an arguably significance (with $p<0.1$) for most datasets. This further confirms the effectiveness of our approach. The performance improvement yielded by the proposed approach is much larger when applying on the vanilla transformer-based text classification model without pretraining. We suspect this may be because the BERT model is already approaching the state-of-the-art performance on each datasets and the dominant attention heads have already been formed in the pretrained BERT model during its pretraining procedure, which reduces the improvement yielded by Scheduled DropHead. Pretraining BERT models with DropHead may alleviate this problem, we leave this for the future work. In addition, we find that the Scheduled DropHead outperforms its counterpart where a constant dropout rate is applied for DropHead. This confirms the effectiveness of the proposed dropout rate schedule.

\subsection{Analysis}

In this section, we perform experiments following that in~\cite{michel2019sixteen} to analyze the relative importance of individual attention heads in transformer model trained on WMT14 en-de dataset with and without the proposed Scheduled DropHead mechanism, and try to prune attention heads to improve its efficiency.

\begin{table}[!ht]
	\centering
		\resizebox{1.\linewidth}{!}{
		\begin{tabular}{lccc}
			\toprule
			\bf Model & \bf Enc-Enc & \bf Enc-Dec & \bf Dec-Dec \\
			\midrule
			\textbf{Transformer} & -0.47 & -1.05 & -0.32 \\
			~-DropHead  & -0.31  &  -0.79 & -0.27 \\
			~-Scheduled DropHead & -0.28  & -0.70  & -0.25 \\
			\textbf{Transformer (more heads)} & -0.41 & -0.92 & -0.29  \\
			~-DropHead  & -0.25 &  -0.61 & -0.24 \\
			~-Scheduled DropHead & \bf -0.21 & \bf -0.54 & \bf -0.20 \\
			\bottomrule
	\end{tabular}}
	\caption{Average variation of the BLEU score after removing the dominant attention head in different models.}
	\label{head1}
\end{table}

\paragraph{Dominant head analysis} The aforementioned experiments on machine translation and text classification datasets demonstrates that our proposed method is able to improve the performance of transformer models. However, as the performance improvement may exclusively come from the better regularization effect yielded by DropHead, it's still unclear whether the DropHead mechanism can prevent the MHA mechanism from being dominated by a few attention heads.  

Therefore, we conduct an experiment to analyze the influence of the dominant head in each transformer layer on the final performance of the transformer model. Specifically, we evaluate the model’s performance
by masking one head at each time and the dominant head is then determined as the one without which the performance of the model degrades the most. We measure the average BLEU score variation of the model's performance after removing the dominant attention head in each layer. Following~\citet{michel2019sixteen}, we report the results when removing encoder-encoder attention heads, decoder-decoder attention heads, and encoder-decoder attention heads separately as they may behave differently.

The results are presented in Table \ref{head1}. We find that the performance degradation when removing the dominant head is significantly reduced when trained with Scheduled DropHead. This suggests that our proposed method can substantially reduce the influence of dominant attention heads and exploit multiple attention heads more effectively. This explains the performance improvement yielded by Scheduled DropHead. In addition, we find that training with Scheduled DropHead can reduce the influence of the dominant heads in the transformer model with more attention heads more effectively, which may explain the improved performance with the transformer model with more attention heads trained with DropHead.

\begin{figure}
    \centering
    \includegraphics[width=1.\linewidth]{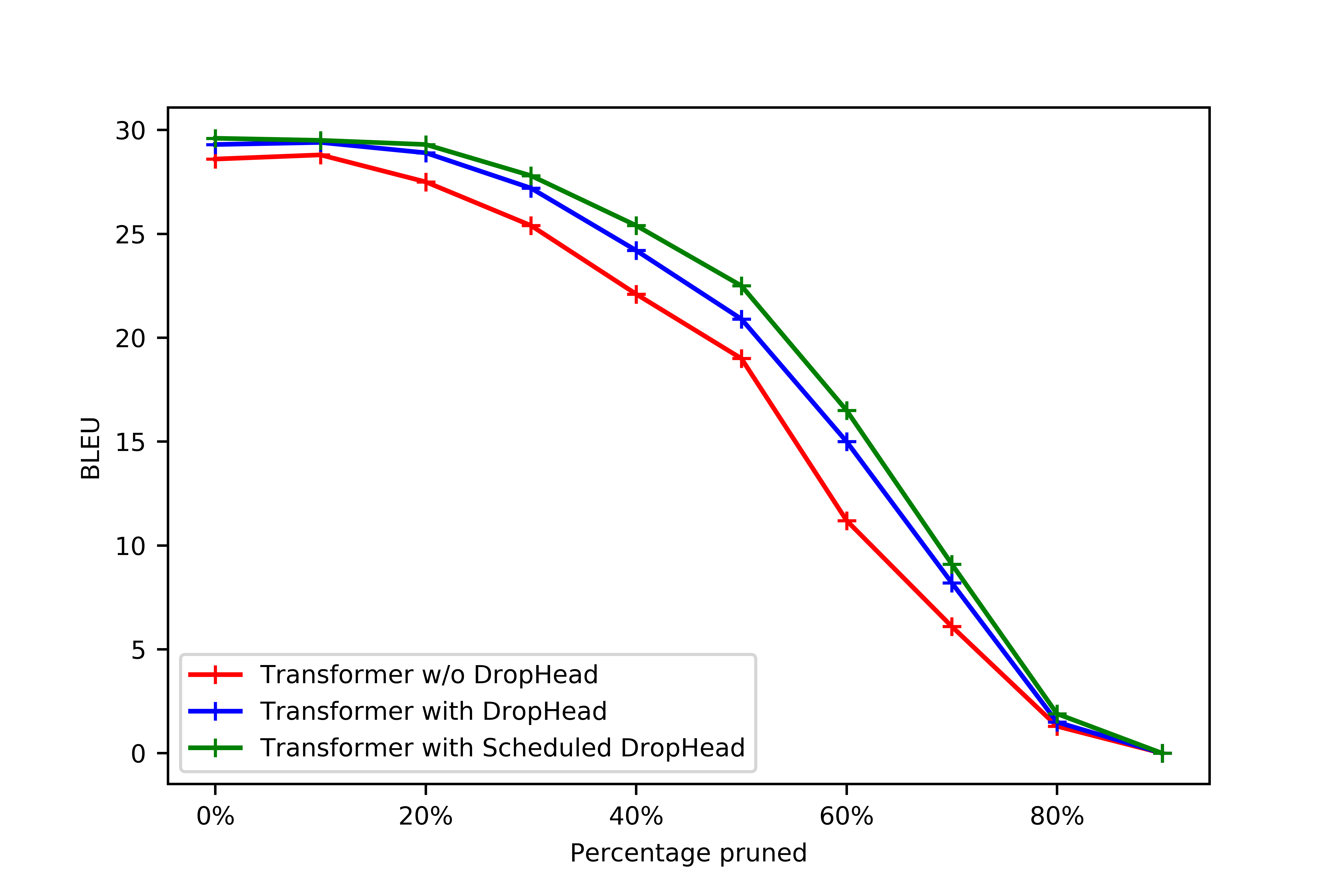}
    \caption{Evolution of BLEU score when attention heads are pruned from different variants of transformer models trained on WMT14 en-de dataset.}
    \label{prune}
\end{figure}

\paragraph{Attention head pruning} Similar to other structured dropout methods~\cite{fan2019reducing,ghiasi2018dropblock}, training with Scheduled DropHead makes the MHA module more robust to perform the task with missing attention heads. This may make it easier to prune attention heads in the trained transformer model at test time to reduce the number of parameters and improve the model's efficiency during inference. To validate this assumption, we perform attention head pruning with estimated head importance score $I_{n}$ as described in~\cite{michel2019sixteen}. Specifically, we prune different portion of the total number of attention heads in the transformer-big model trained with and without the DropHead mechanism and the proposed dropout rate schedule, and report the evolution of BLEU score in Table \ref{prune}. We can see that both the DropHead mechanism and the dropout rate scheduler can increase the model's robustness to attention head pruning. This allows us to prune more attention heads in a transformer model to reduce the parameters in the model and improve the inference efficiency without significantly sacrificing the model's performance.

\subsection{Ablation study}

We also conduct an ablation study to better understand DropHead and the dropout rate schedule. We analyze the relationship between standard Dropout and Scheduled DropHead; the effect of Scheduled DropHead on different types of attention heads; and the effect of different variants of dropout rate schedules. The experiments are conducted on the machine translation task with WMT14 en-de dataset.

\paragraph{Attention Dropout vs Scheduled DropHead} To analyze the relationship between Scheduled DropHead and the standard Dropout applied to the self-attention layers in the transformer model, we present the performance of transformer models trained by applying standard Dropout, Scheduled DropHead, and the combination of them on the MHA modules in Table \ref{ablation1}. We can see that applying standard dropout on the MHA modules only marginally improve the model's performance. In contrast, the proposed Scheduled DropHead can yield significant improvement upon the baseline model and the model trained with standard dropout. We also find that the combination of the standard dropout and DropHead does not yield further improvement. This suggests that Scheduled DropHead is more effective than the standard Dropout for the MHA mechanism.

\begin{table}[!ht]
	\centering
	\scalebox{1.0}{
		\begin{tabular}{ll}
			\toprule
			\bf Model & \bf BLEU  \\
			\midrule
			\textbf{Transformer} &  28.7 \\
			~-+ Attention Dropout & 28.7  \\
			~-+ Scheduled DropHead & \bf 29.4 \\
			~-+ Combination & \bf 29.3 \\
			\bottomrule
	\end{tabular}}
	\caption{BLEU score of transformer-big models trained with different regularization methods.}
	\label{ablation1}
\end{table}

\paragraph{Influence on different types of attention heads} We also analyze the effect of Scheduled DropHead on different types of attention heads. Specifically, we apply Scheduled DropHead separately on encoder-encoder attention, encoder-decoder attention, and decoder-decoder attention modules. The results are presented in Table \ref{ablation2}. We can see that the proposed approach is more effective when used on encoder-encoder self-attention and encoder-decoder attention modules, which is consistent with the finding of~\cite{michel2019sixteen}.

\begin{table}[!ht]
	\centering
	\scalebox{1.0}{
		\begin{tabular}{lllll}
			\toprule
			\bf Model & \bf BLEU  \\
			\midrule
			\textbf{Transformer} &   \\
			~-w/o Scheduled DropHead  & 28.6  \\
			~-Enc-enc Scheduled DropHead & 28.9 \\
			~-Enc-dec Scheduled DropHead & 29.1 \\
			~-Dec-dec Scheduled DropHead & 28.7 \\
			~-Scheduled DropHead & \bf 29.4 \\
			\bottomrule
	\end{tabular}}
	\caption{BLEU score of transformer-big models trained with Scheduled DropHead applied on different MHA modules.}
	\label{ablation2}
\end{table}

\begin{table}[!ht]
	\centering
	\scalebox{1.0}{
		\begin{tabular}{lllll}
			\toprule
			\bf Model & \bf BLEU  \\
			\midrule
			\textbf{Transformer} &   \\
			~-DropHead (constant dropout rate) & 29.2 \\
			~-Curriculum DropHead & 29.0 \\
			~-Anti-curriculum DropHead & 28.7 \\
			~-Scheduled DropHead & \bf 29.4 \\
			\bottomrule
	\end{tabular}}
	\caption{BLEU score of transformer-big models trained with Scheduled DropHead with different dropout rate schedules.}
	\label{ablation3}
\end{table}

\paragraph{Comparison between different dropout rate schedules} We then analyze the influence of different dropout rate schedules by comparing the proposed dropout schedule with two variants. As illustrated in Figure \ref{dropoutrate}, the first variant (curriculum dropout schdule) is that used in ScheduledDropPath~\cite{zoph2018learning} which linearly increases the dropout rate from 0 to the pre-defined dropout rate. The second (anti-curriculum dropout schdule) is the opposite of the former schedule which linearly decreases the dropout rate from the pre-defined dropout rate to 0. The result is shown in Table \ref{ablation3}. We can see that both curriculum dropout rate scheduler and anti-curriculum dropout schedule fails to improve the performance of the vanilla DropHead where the dropout rate is constant, which demonstrates the effectiveness of the proposed dropout rate schedule.

\section{Related work}

Since its introduction, dropout~\cite{srivastava2014dropout} has inspired a number of regularization methods for neural networks, including DropConnect~\cite{wan2013regularization}, Maxout~\cite{goodfellow2013maxout}, Spatial Dropout~\cite{tompson2015efficient}, StochasticDepth~\cite{huang2016deep}, DropPath~\cite{larsson2016fractalnet}, DropBlock~\cite{ghiasi2018dropblock}, LayerDrop~\cite{fan2019reducing}, etc. While standard dropout works well for fully-connected neural networks, structured dropout~\cite{huang2016deep,larsson2016fractalnet,ghiasi2018dropblock} that randomly drops an entire channel or input block appears to work better in convolutional neural networks. For RNNs, Variational Dropout~\cite{gal2016theoretically}, ZoneOut~\cite{krueger2016zoneout}, and Word Embedding Dropout in both word-level~\cite{gal2016theoretically} and embedding-level~\cite{zhou2019bert} are the most widely used methods. More recently, there exists work investigates employing standard Dropout on attention layers~\cite{zehui2019dropattention}, or applying StochasticDepth on transformer layers~\cite{fan2019reducing}. However, to the best of our knowledge, regularization approaches specifically designed for the MHA mechasim in the transformer architecture are currently under-explored. 

DropHead is also closely related to Spatial Dropout~\cite{tompson2015efficient}, which is a structured dropout method for convolutional neural networks that randomly drops an entire channel from a feature map. The proposed dropout rate scheduler is related to the Curriculum Dropout and the dropout rate scheduler in ScheduledDropPath and DropBlock. Finally, our work is also inspired by recent work analyzing the role of one attention head in the MHA mechanism~\cite{michel2019sixteen,voita2019analyzing} which found that most attention heads in MHA fail to contribute to the performance of the transformer model and can be effectively pruned at test time.


%

\section{Conclusion}

In this work, we introduce Scheduled DropHead, a simple regularization approach for training transformer models. DropHead is a form of structured dropout that drops an entire attention head in the multi-head attention mechanism, which prevents MHA from being dominated by a few heads, reduces excessive co-adaptation between attention heads, and facilitates attention head pruning. We also propose a specific dropout rate scheduler that is motivated by the training dynamic of the MHA mechanism to improve the performance of the vanilla DropHead. Our experiments on machine translation and text classification benchmarks demonstrate that the DropHead mechanism and the dropout rate scheduler can effectively improve the performance of a competitive transformer model. We also find that the proposed approach can reduce the influence of the dominant attention head and improve the model's robustness to attention head pruning. 

For future work, we plan to apply DropHead on pretraining language models such as BERT as well as other natural language generation tasks such as text summarization and dialogue systems.

\section*{Acknowledgments}
We thank the anonymous reviewers for their valuable comments.

\bibliography{acl2019}

\begin{thebibliography}{32}
\expandafter\ifx\csname natexlab\endcsname\relax\def\natexlab#1{#1}\fi

\bibitem[{Ahmed et~al.(2017)Ahmed, Keskar, and Socher}]{ahmed2017weighted}
Karim Ahmed, Nitish~Shirish Keskar, and Richard Socher. 2017.
\newblock Weighted transformer network for machine translation.
\newblock \emph{arXiv preprint arXiv:1711.02132}.

\bibitem[{Bahdanau et~al.(2014)Bahdanau, Cho, and Bengio}]{bahdanau2014neural}
Dzmitry Bahdanau, Kyunghyun Cho, and Yoshua Bengio. 2014.
\newblock Neural machine translation by jointly learning to align and
  translate.
\newblock \emph{arXiv preprint arXiv:1409.0473}.

\bibitem[{Bowman et~al.(2015)Bowman, Angeli, Potts, and
  Manning}]{bowman2015large}
Samuel~R Bowman, Gabor Angeli, Christopher Potts, and Christopher~D Manning.
  2015.
\newblock A large annotated corpus for learning natural language inference.
\newblock \emph{arXiv preprint arXiv:1508.05326}.

\bibitem[{Conneau et~al.(2017)Conneau, Kiela, Schwenk, Barrault, and
  Bordes}]{conneau2017supervised}
Alexis Conneau, Douwe Kiela, Holger Schwenk, Loic Barrault, and Antoine Bordes.
  2017.
\newblock Supervised learning of universal sentence representations from
  natural language inference data.
\newblock \emph{arXiv preprint arXiv:1705.02364}.

\bibitem[{Devlin et~al.(2018)Devlin, Chang, Lee, and
  Toutanova}]{devlin2018bert}
Jacob Devlin, Ming-Wei Chang, Kenton Lee, and Kristina Toutanova. 2018.
\newblock Bert: Pre-training of deep bidirectional transformers for language
  understanding.
\newblock \emph{arXiv preprint arXiv:1810.04805}.

\bibitem[{Fan et~al.(2019)Fan, Grave, and Joulin}]{fan2019reducing}
Angela Fan, Edouard Grave, and Armand Joulin. 2019.
\newblock Reducing transformer depth on demand with structured dropout.
\newblock \emph{arXiv preprint arXiv:1909.11556}.

\bibitem[{Gal and Ghahramani(2016)}]{gal2016theoretically}
Yarin Gal and Zoubin Ghahramani. 2016.
\newblock A theoretically grounded application of dropout in recurrent neural
  networks.
\newblock In \emph{Advances in neural information processing systems}, pages
  1019--1027.

\bibitem[{Ghiasi et~al.(2018)Ghiasi, Lin, and Le}]{ghiasi2018dropblock}
Golnaz Ghiasi, Tsung-Yi Lin, and Quoc~V Le. 2018.
\newblock Dropblock: A regularization method for convolutional networks.
\newblock In \emph{Advances in Neural Information Processing Systems}, pages
  10727--10737.

\bibitem[{Goodfellow et~al.(2013)Goodfellow, Warde-Farley, Mirza, Courville,
  and Bengio}]{goodfellow2013maxout}
Ian~J Goodfellow, David Warde-Farley, Mehdi Mirza, Aaron Courville, and Yoshua
  Bengio. 2013.
\newblock Maxout networks.
\newblock \emph{arXiv preprint arXiv:1302.4389}.

\bibitem[{He et~al.(2018)He, Tan, Xia, He, Qin, Chen, and Liu}]{he2018layer}
Tianyu He, Xu~Tan, Yingce Xia, Di~He, Tao Qin, Zhibo Chen, and Tie-Yan Liu.
  2018.
\newblock Layer-wise coordination between encoder and decoder for neural
  machine translation.
\newblock In \emph{Advances in Neural Information Processing Systems}, pages
  7944--7954.

\bibitem[{Howard and Ruder(2018)}]{howard2018universal}
Jeremy Howard and Sebastian Ruder. 2018.
\newblock Universal language model fine-tuning for text classification.
\newblock \emph{arXiv preprint arXiv:1801.06146}.

\bibitem[{Huang et~al.(2016)Huang, Sun, Liu, Sedra, and
  Weinberger}]{huang2016deep}
Gao Huang, Yu~Sun, Zhuang Liu, Daniel Sedra, and Kilian~Q Weinberger. 2016.
\newblock Deep networks with stochastic depth.
\newblock In \emph{European conference on computer vision}, pages 646--661.
  Springer.

\bibitem[{Johnson and Zhang(2017)}]{johnson2017deep}
Rie Johnson and Tong Zhang. 2017.
\newblock Deep pyramid convolutional neural networks for text categorization.
\newblock In \emph{Proceedings of the 55th Annual Meeting of the Association
  for Computational Linguistics (Volume 1: Long Papers)}, pages 562--570.

\bibitem[{Krueger et~al.(2016)Krueger, Maharaj, Kram{\'a}r, Pezeshki, Ballas,
  Ke, Goyal, Bengio, Courville, and Pal}]{krueger2016zoneout}
David Krueger, Tegan Maharaj, J{\'a}nos Kram{\'a}r, Mohammad Pezeshki, Nicolas
  Ballas, Nan~Rosemary Ke, Anirudh Goyal, Yoshua Bengio, Aaron Courville, and
  Chris Pal. 2016.
\newblock Zoneout: Regularizing rnns by randomly preserving hidden activations.
\newblock \emph{arXiv preprint arXiv:1606.01305}.

\bibitem[{Larsson et~al.(2016)Larsson, Maire, and
  Shakhnarovich}]{larsson2016fractalnet}
Gustav Larsson, Michael Maire, and Gregory Shakhnarovich. 2016.
\newblock Fractalnet: Ultra-deep neural networks without residuals.
\newblock \emph{arXiv preprint arXiv:1605.07648}.

\bibitem[{Luong et~al.(2015)Luong, Pham, and Manning}]{luong2015effective}
Minh-Thang Luong, Hieu Pham, and Christopher~D Manning. 2015.
\newblock Effective approaches to attention-based neural machine translation.
\newblock \emph{arXiv preprint arXiv:1508.04025}.

\bibitem[{Maas et~al.(2011)Maas, Daly, Pham, Huang, Ng, and
  Potts}]{maas2011learning}
Andrew~L Maas, Raymond~E Daly, Peter~T Pham, Dan Huang, Andrew~Y Ng, and
  Christopher Potts. 2011.
\newblock Learning word vectors for sentiment analysis.
\newblock In \emph{Proceedings of the 49th annual meeting of the association
  for computational linguistics: Human language technologies-volume 1}, pages
  142--150. Association for Computational Linguistics.

\bibitem[{Michel et~al.(2019)Michel, Levy, and Neubig}]{michel2019sixteen}
Paul Michel, Omer Levy, and Graham Neubig. 2019.
\newblock Are sixteen heads really better than one?
\newblock \emph{Advances in neural information processing systems (To appear)}.

\bibitem[{Morerio et~al.(2017)Morerio, Cavazza, Volpi, Vidal, and
  Murino}]{morerio2017curriculum}
Pietro Morerio, Jacopo Cavazza, Riccardo Volpi, Ren{\'e} Vidal, and Vittorio
  Murino. 2017.
\newblock Curriculum dropout.
\newblock In \emph{Proceedings of the IEEE International Conference on Computer
  Vision}, pages 3544--3552.

\bibitem[{Sennrich et~al.(2015)Sennrich, Haddow, and
  Birch}]{sennrich2015neural}
Rico Sennrich, Barry Haddow, and Alexandra Birch. 2015.
\newblock Neural machine translation of rare words with subword units.
\newblock \emph{arXiv preprint arXiv:1508.07909}.

\bibitem[{Srivastava et~al.(2014)Srivastava, Hinton, Krizhevsky, Sutskever, and
  Salakhutdinov}]{srivastava2014dropout}
Nitish Srivastava, Geoffrey Hinton, Alex Krizhevsky, Ilya Sutskever, and Ruslan
  Salakhutdinov. 2014.
\newblock Dropout: a simple way to prevent neural networks from overfitting.
\newblock \emph{The journal of machine learning research}, 15(1):1929--1958.

\bibitem[{Sun et~al.(2019)Sun, Qiu, Xu, and Huang}]{sun2019fine}
Chi Sun, Xipeng Qiu, Yige Xu, and Xuanjing Huang. 2019.
\newblock How to fine-tune bert for text classification?
\newblock In \emph{China National Conference on Chinese Computational
  Linguistics}, pages 194--206. Springer.

\bibitem[{Tompson et~al.(2015)Tompson, Goroshin, Jain, LeCun, and
  Bregler}]{tompson2015efficient}
Jonathan Tompson, Ross Goroshin, Arjun Jain, Yann LeCun, and Christoph Bregler.
  2015.
\newblock Efficient object localization using convolutional networks.
\newblock In \emph{Proceedings of the IEEE Conference on Computer Vision and
  Pattern Recognition}, pages 648--656.

\bibitem[{Vaswani et~al.(2017)Vaswani, Shazeer, Parmar, Uszkoreit, Jones,
  Gomez, Kaiser, and Polosukhin}]{vaswani2017attention}
Ashish Vaswani, Noam Shazeer, Niki Parmar, Jakob Uszkoreit, Llion Jones,
  Aidan~N Gomez, {\L}ukasz Kaiser, and Illia Polosukhin. 2017.
\newblock Attention is all you need.
\newblock In \emph{Advances in neural information processing systems}, pages
  5998--6008.

\bibitem[{Voita et~al.(2019)Voita, Talbot, Moiseev, Sennrich, and
  Titov}]{voita2019analyzing}
Elena Voita, David Talbot, Fedor Moiseev, Rico Sennrich, and Ivan Titov. 2019.
\newblock Analyzing multi-head self-attention: Specialized heads do the heavy
  lifting, the rest can be pruned.
\newblock \emph{Proceedings of the 57th Annual Meeting of the Association for
  Computational Linguistics}.

\bibitem[{Voorhees and Tice(1999)}]{voorhees1999trec}
Ellen~M Voorhees and Dawn~M Tice. 1999.
\newblock The trec-8 question answering track evaluation.
\newblock In \emph{TREC}, volume 1999, page~82. Citeseer.

\bibitem[{Wan et~al.(2013)Wan, Zeiler, Zhang, Le~Cun, and
  Fergus}]{wan2013regularization}
Li~Wan, Matthew Zeiler, Sixin Zhang, Yann Le~Cun, and Rob Fergus. 2013.
\newblock Regularization of neural networks using dropconnect.
\newblock In \emph{International conference on machine learning}, pages
  1058--1066.

\bibitem[{Xia et~al.(2019)Xia, He, Tan, Tian, He, and Qin}]{xia2019tied}
Yingce Xia, Tianyu He, Xu~Tan, Fei Tian, Di~He, and Tao Qin. 2019.
\newblock Tied transformers: Neural machine translation with shared encoder and
  decoder.
\newblock In \emph{Proceedings of the AAAI Conference on Artificial
  Intelligence}, volume~33, pages 5466--5473.

\bibitem[{Zehui et~al.(2019)Zehui, Liu, Huang, Fu, Chen, Qiu, and
  Huang}]{zehui2019dropattention}
Lin Zehui, Pengfei Liu, Luyao Huang, Jie Fu, Junkun Chen, Xipeng Qiu, and
  Xuanjing Huang. 2019.
\newblock Dropattention: A regularization method for fully-connected
  self-attention networks.
\newblock \emph{arXiv preprint arXiv:1907.11065}.

\bibitem[{Zhang et~al.(2015)Zhang, Zhao, and LeCun}]{zhang2015character}
Xiang Zhang, Junbo Zhao, and Yann LeCun. 2015.
\newblock Character-level convolutional networks for text classification.
\newblock In \emph{Advances in neural information processing systems}, pages
  649--657.

\bibitem[{Zhou et~al.(2019)Zhou, Ge, Xu, Wei, and Zhou}]{zhou2019bert}
Wangchunshu Zhou, Tao Ge, Ke~Xu, Furu Wei, and Ming Zhou. 2019.
\newblock Bert-based lexical substitution.
\newblock In \emph{Proceedings of the 57th Annual Meeting of the Association
  for Computational Linguistics}, pages 3368--3373.

\bibitem[{Zoph et~al.(2018)Zoph, Vasudevan, Shlens, and Le}]{zoph2018learning}
Barret Zoph, Vijay Vasudevan, Jonathon Shlens, and Quoc~V Le. 2018.
\newblock Learning transferable architectures for scalable image recognition.
\newblock In \emph{Proceedings of the IEEE conference on computer vision and
  pattern recognition}, pages 8697--8710.

\end{thebibliography}
\bibliographystyle{acl_natbib}

\end{document}